# Unsupervised Local Plasticity in a Multi-Frequency VisNet Hierarchy

Learning Visual Representations without Labels or Backpropagation

**Mehdi Fatan Serj  C. Alejandro Parraga  Xavier Otazu**

Computer Vision Centre (CVC), Universitat Autònoma de Barcelona, Spain

mfatan@cvc.uab.cat

**Abstract**

We present an unsupervised visual feature learning system built from local plasticity rules: no labels, no backpropagation, and no global error signals are used to update representational weights. The architecture is a VisNet-based hierarchy with opponent-colour inputs, multi-frequency Gabor and wavelet streams, competitive normalisation with lateral inhibition, a saliency-gated side branch, an associative memory module, and a two-pass feedback loop. Every learnable representational weight is updated by unsupervised local plasticity rules operating continuously on unlabeled image streams throughout all 300 training epochs.

To evaluate the quality of the learned representation, we attach a minimal linear classifier and measure classification accuracy. This readout is the *only* component that uses labels or gradient descent; it serves as a measuring instrument, not as part of the learning system.

On CIFAR-10, the system achieves 80.1% (+9.1 pp over a Hebbian-only baseline); on CIFAR-100, 47.6% (+9.4 pp). A controlled ablation study identifies anti-Hebbian decorrelation, free-energy plasticity, and associative memory as the three most important contributors, with strong super-additive interactions. Evaluating only the fixed architectural priors with no learning yields 61.4% on CIFAR-10, confirming that the plasticity rules—not the hand-crafted front end alone—are responsible for the majority of the gain.

A fresh-probe control confirms that a new linear classifier trained from scratch on the final frozen representation produces accuracy within 0.3 pp of the co-trained probe. A nearest-class-mean readout achieves 78.3% with no gradient computation in the readout, further confirming intrinsic feature quality.

The residual gap to a parameter-matched CNN trained end-to-end by backpropagation is 5.7 pp on CIFAR-10 and 7.5 pp on CIFAR-100. We do not claim to close this gap. We show that structured local plasticity can learn useful visual representations from raw, unlabeled pixels without gradient-based credit assignment in the representation—a capability relevant to settings where labels are scarce or absent, and potentially to future low-power hardware implementations.

## 1. Introduction

Modern deep learning has converged on a common training paradigm: collect a large labeled dataset, define a differentiable objective, and propagate error gradients end-to-end through the network. This paradigm is extremely effective, but it depends on three assumptions that are often difficult to satisfy in natural or embedded settings:

- labels are available at training time,
- a global backward pass can be computed,
- training can revisit stored examples many times.

Biological visual systems do not appear to rely on this combination. The primate ventral stream builds rich visual representations from continuous, unlabeled sensory input using local synaptic plasticity, without any known mechanism equivalent to backpropagation through arbitrary depth [23, 34]. This motivates a basic question:

> *How good a visual representation can be learned from unlabeled images using only local plasticity rules, with no labels and no gradient-based credit assignment in the representation?*

**What this paper studies.** We construct a hierarchical visual system in which **every learnable representational weight is updated by unsupervised local plasticity**. The model never sees class labels while learning its internal representation. After this unsupervised learning process, we evaluate the resulting features using a minimal linear classifier. This evaluation protocol is standard in unsupervised and self-supervised representation learning [3]: the classifier is a probe, not part of the scientific claim about how the representation is acquired.

**Design philosophy.** The architecture borrows computational motifs from visual neuroscience—opponent-colour inputs, Gabor-like orientation-selective filters, divisive normalisation, anti-Hebbian decorrelation, associative memory, top-down feedback—because these are rigorous computational primitives with independent empirical support, not because we are modelling cortex. We use labels like "side branch", "main pathway", and "associative memory" as functional shorthand; all biological framing should be read as functional rather than mechanistic.

**What this paper does not claim.** We do not claim to reproduce cortical biology, to outperform backpropagation, or that every computation in the architecture is pointwise local in a strict physical sense. The claim is narrower: useful visual representations can be learned from raw pixels using local plasticity rules, without label-derived error signals and without a backward pass through the representation.

**Evaluation scope.** The full component ablation, pairwise interaction analysis, side-branch controls, and prototype readout are conducted on CIFAR-10. CIFAR-100 is included as a cross-dataset robustness check with headline numbers only.

**Note on hyperparameter selection.** While no labels enter the representation learning process itself, hyperparameter choices (batch size, rule configurations, number of memory slots) were guided by linear-probe accuracy on a validation split. This is standard practice in unsupervised representation learning [3], but we acknowledge that the overall pipeline is not fully label-free: labels influence architectural decisions even though they do not influence representational weights.

**Relationship to prior work.** This paper extends our earlier VisNet-based framework [12], which added opponent-colour coding, multi-frequency streams, and multiple local plasticity rules to the classic VisNet architecture. The present work adds a saliency-gated side branch, associative memory, feedback, a stronger evaluation protocol, and explicit controls that separate architectural priors from learned plasticity effects.

**Contributions.**

1. **A fully unsupervised representation learner.** All learnable representational weights are updated by local plasticity on unlabeled images throughout all 300 training epochs; labels are used only for the evaluation readout.
2. **Architectural-prior isolation.** An epoch-0 / no-learning baseline isolates the contribution of fixed front-end structure from the contribution of local plasticity.
3. **Probe-independence verification.** A fresh-probe control shows that a newly trained probe on the final frozen representation achieves accuracy within 0.3 pp of the co-trained probe.
4. **Component ranking and interactions on CIFAR-10.** Anti-Hebbian decorrelation ($-5.5$ pp), free-energy plasticity ($-4.2$ pp), and associative memory ($-3.2$ pp) are the three largest contributors, with strong super-additive interactions.
5. **Gradient-importance finding.** A no-stop-gradient control shows that permitting gradients into the representation adds $+3.3$ pp on CIFAR-10 and $+4.2$ pp on CIFAR-100.
6. **Gradient-free readout.** A nearest-class-mean readout achieves 78.3% on the frozen learned features, confirming that the representation supports strong classification without any gradient-based readout training.

## 2. Related Work

**Local learning rules.** The biological implausibility of backpropagation has motivated local alternatives including Oja's rule [28], Equilibrium Propagation [35], Target Propagation [22], Forward-Forward [16], PEPITA [7], and predictive coding [14, 31]. Our design restricts representational updates to pre-/post-synaptic unsupervised rules; the only gradient-trained component is the final linear readout, explicitly isolated by a `stop-gradient` barrier.

**Unsupervised representation learning.** Contrastive and self-supervised methods such as SimCLR [3] achieve strong performance without labels, but still rely on end-to-end backpropagation through a deep network. Our work differs in placing the representation itself outside the gradient-training paradigm.

**VisNet and its extensions.** VisNet [32] is a four-layer hierarchical model of the ventral visual stream using Oja-style Hebbian learning with competitive normalisation. Fatan Serj et al. [12] extended VisNet with opponent-colour inputs, multi-frequency Gabor and Haar wavelet streams, and multiple local plasticity rules. We build on that foundation by adding a saliency-gated side branch, associative memory, and a two-pass feedback loop.

**Early vision.** Opponent-colour coding [8, 36] and Gabor models of V1 simple cells [5, 18, 20] are well established. Haar wavelets [25] provide complementary scale-aligned chromatic features.

**Competitive learning and normalisation.** Anti-Hebbian decorrelation [13] and divisive normalisation [2, 24] are canonical neural computations for shaping sparse representations. Our competitive stream architecture incorporates both [12].

**Saliency and auxiliary pathways.** Computational saliency models [19] capture bottom-up attentional biases. We use a static saliency map to gate a secondary pathway. Unlike motion-processing models [1], our side branch operates on single frames.

**Associative memory.** Modern Hopfield networks achieve exponential capacity via softmax attention [30]. We use this framework as a memory module whose keys and values are updated by local Hebbian rules, not by gradient descent.

**Top-down feedback.** The feedback signal is loosely inspired by context-dependent gating [26] and predictive coding [31]. It provides a within-sample mechanism for binding distributed features; the biology is inspirational rather than mechanistic.

**Neuromorphic computing.** Local plasticity rules are a natural fit for event-driven hardware [6, 33]. However, there is a substantial gap between the current floating-point implementation and actual neuromorphic deployment (see §6.7).

# 3. Architecture

## 3.1. Overview

The architecture extends the VisNet framework [12, 32] with three new components: a saliency-gated side branch (§3.6), an associative memory module (§3.8), and a two-pass feedback loop (§3.9). The multi-frequency front end, competitive streams, and local plasticity rules are inherited from [12].

Each input is processed in **two within-sample passes**. Pass 1 is feedforward: opponent-colour input → Gabor/Haar front end → four-layer competitive hierarchy → saliency-gated side branch → fusion → memory module. Pass 2 broadcasts the memory output as feedback; the hierarchy re-runs on modulated inputs. A hard `stop-gradient` barrier then separates the resulting representation from the final linear classifier.

All local plasticity rules remain active throughout the full 300 training epochs; no representational component is frozen early.

**Definition 1** (Gradient isolation)**.** *Let* $\mathbf{z}_{\text{final}} \in \mathbb{R}^{512}$ *denote the representation after Pass 2. The input to the classifier is* $\bar{\mathbf{z}} = \text{sg}(\mathbf{z}_{\text{final}})$*, where:*

$$\text{sg}(\mathbf{z}) = \mathbf{z} \text{ (forward)}, \qquad \frac{\partial \, \text{sg}(\mathbf{z})}{\partial \, \mathbf{z}} = \mathbf{0} \text{ (backward)}. \tag{1}$$

*This guarantees* $\partial \mathcal{L}_{\text{CE}} / \partial \boldsymbol{\theta}_{\text{repr}} = \mathbf{0}$ *for every representational parameter.*

## 3.2. Opponent-Colour Input and Wavelet Front End

Following [12], the opponent transform converts RGB into cardinal colour directions [8, 36]:

$$\mathbf{X}_t^{\text{Opp}} = (L{+}M,\ L{-}M,\ S{-}(L{+}M)) \in \mathbb{R}^{H \times W \times 3}. \tag{2}$$

Table 1: System at a glance.

| Property | Value |
|---|---|
| Hierarchy depth | 4 layers (L1–L4) |
| Number of frequency streams | 7 (Gabor) + 1 (Haar) |
| Side-branch layers | 2 |
| Hopfield memory slots | 96 |
| Input resolution | $32 \times 32$ (CIFAR native) |
| Representation dimensionality | 512 (§3.12) |
| Total representational params | ≈1.2M (§3.12) |
| Datasets | CIFAR-10 (primary), CIFAR-100 (robustness check) |
| Passes per sample | 2 (feedforward + feedback) |
| Plasticity duration | All 300 epochs (continuous) |
| Data augmentation | None (§4) |
| Plasticity rules (CIFAR-10 reduced / full) | 4 / 7 |
| Plasticity rules (CIFAR-100 reduced / full) | 5 / 8 (§5.11) |
| Classifier | Single linear layer (stop-gradient isolated) |

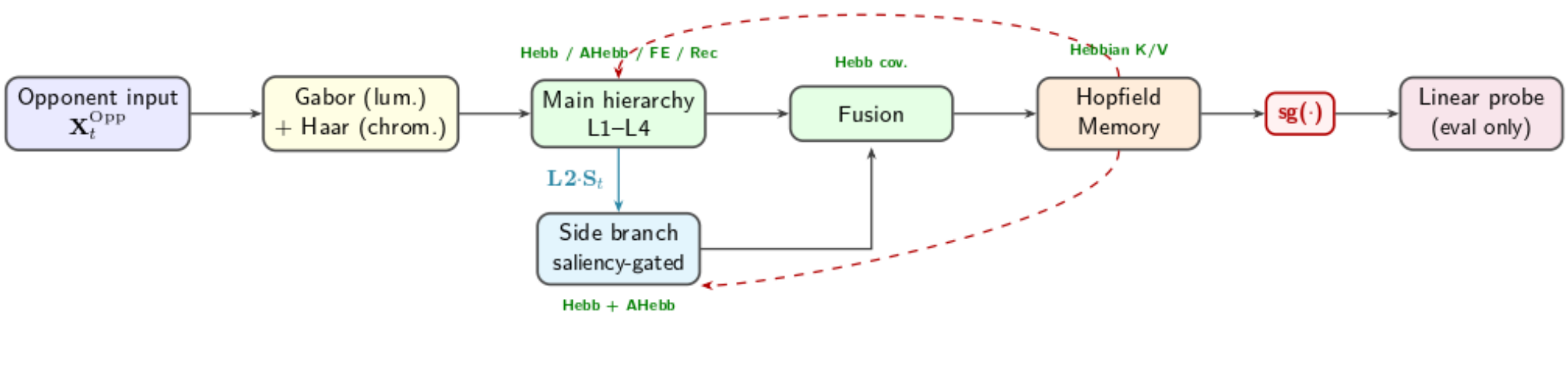

**Green modules:** unsupervised local plasticity. **Purple:** evaluation probe only (labels + gradient).
**sg(·)**: stop-gradient barrier. Red dashed: feedback (Pass 2).

Figure 1: Architecture overview. All learnable modules to the left of the stop-gradient barrier are trained by unsupervised local plasticity. The linear probe evaluates the learned representation.

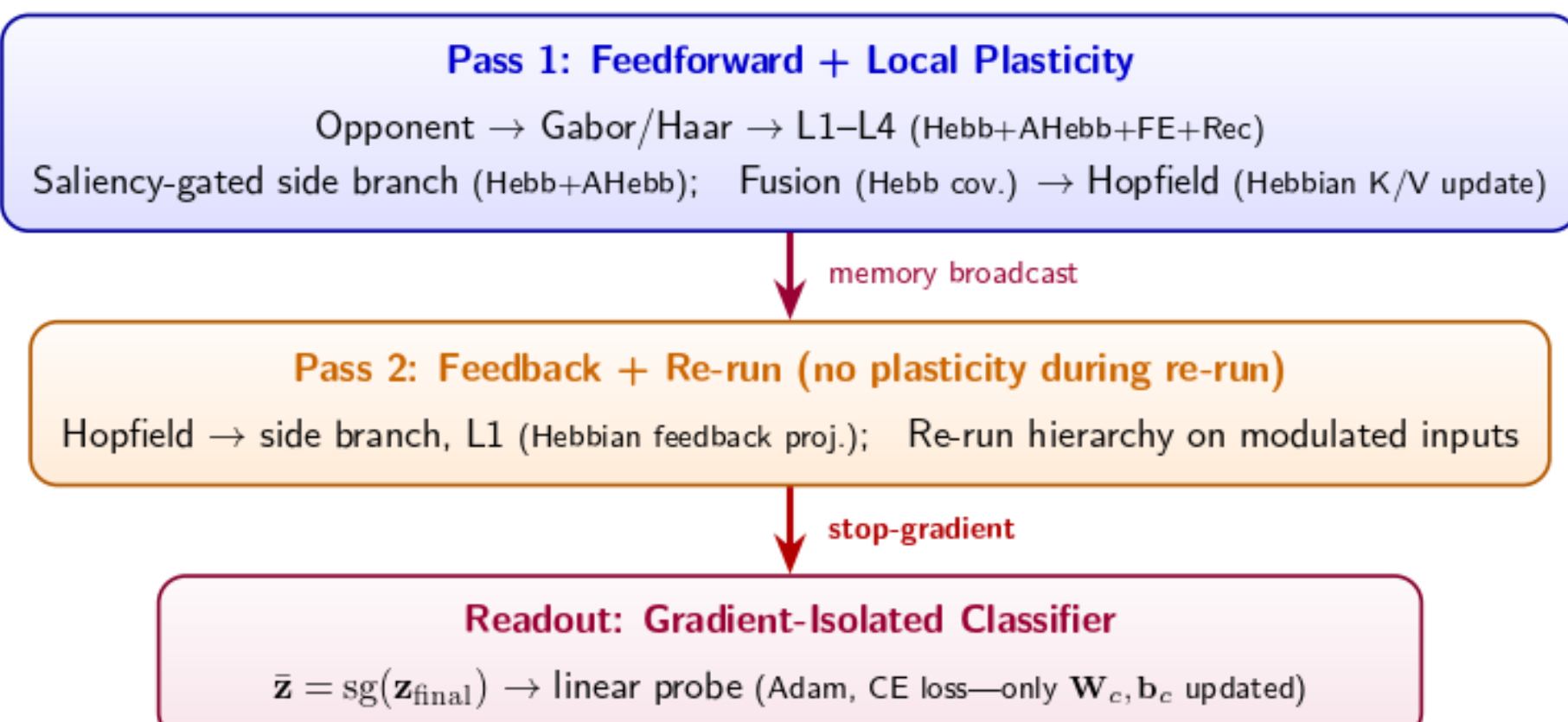

Figure 2: Two-pass timeline with gradient isolation. Local plasticity rules update representational weights during Pass 1. Pass 2 applies feedback modulation with no plasticity during the re-run. The stop-gradient barrier severs the graph before the probe.

**Luminance Gabor bank.**

$$g_{f,\theta,\varphi}(x,y)=\exp\Big(-\frac{x_\theta^2+y_\theta^2}{2\sigma_f^2}\Big)\cos(2\pi f x_\theta+\varphi),\qquad \sigma_f=0.56/f, \tag{3}$$

with $N_f$=7 frequencies, $n_\theta$=7 orientations, and two phases. Phase-invariant energy:

$$\mathbf{G}_{s,t}=\sqrt{(g_s^{\mathrm{re}}*L_t)^2+(g_s^{\mathrm{im}}*L_t)^2+\epsilon}. \tag{4}$$

**Chromatic Haar wavelet.**

$$\mathbf{H}_t=\mathrm{Haar}_2(\mathbf{X}_t^{\mathrm{Opp}})\in\mathbb{R}^{H\times W\times 12}. \tag{5}$$

### 3.3. Competitive Streams with Lateral Inhibition

The four-layer competitive hierarchy follows [12]:

$$y_i^{(s,l)}=\frac{\phi(\tilde{h}_i^{(s,l)}-\alpha\sum_{j\neq i}(L_{ij}^{(l)}\odot\mathcal{M}_{ij}^{(l)})y_j^{(s,l)})}{\sqrt{\alpha_{\mathrm{div}}+\beta_{\mathrm{div}}(w_g\bar{a}^{(l)}+w_\ell a_{\mathrm{local}}^{(l)}\circledast G_{\mathrm{loc}})^2}}, \tag{6}$$

where $\phi(\cdot)$ denotes the pointwise ReLU activation and $\sigma(\cdot)$ denotes the logistic sigmoid used for gating.

### 3.4. Homeostatic Mechanisms

$$g_i\leftarrow(1-\eta_g)g_i+\eta_g\bar{y}_i,\qquad \tilde{h}_i^{(s,l)}=h_i^{(s,l)}-\kappa_g(g_i-\mathrm{clip}(g_i,[0.75,1.25])). \tag{7}$$

### 3.5. Local Plasticity Rules

All learnable representational weights are updated by local, unsupervised rules:

$$\Delta\mathbf{W}^{(s,l)}=\rho_i\big[\alpha_H\Delta^{\mathrm{Hebb}}+\alpha_A\Delta^{\mathrm{AHebb}}+\lambda_F\Delta^{\mathrm{FE}}+\alpha_R\Delta^{\mathrm{Rec}}\big], \tag{8}$$

where $\rho_i$ is a per-neuron plasticity gain clipped to $[0.5,1.5]$.

**Hebbian [15].** $\Delta_{ij}^{\mathrm{Hebb}}=B^{-1}\sum_b y_i^{(b)}x_j^{(b)}-\delta_H W_{ij}$.

**Anti-Hebbian [13].** $\Delta_{ij}^{\mathrm{AHebb}}=-B^{-1}\sum_b y_i^{(b)}y_j^{(b)}$, for $i{\neq}j$. This is the central mechanism through which the hierarchy avoids redundancy.

**Free-energy [14].** $\Delta_{ij}^{\mathrm{FE}}=y_i(x_j-\hat{x}_j)-\lambda_F W_{ij}$, where $\hat{x}_j=\sum_k W_{kj}y_k$ is a locally reconstructed input computed from the *same layer's* weights and activations. The gradient implicit in $\partial\hat{x}_j/\partial W_{ij}$ is a single-layer local gradient, not a multi-layer backward pass.

**Recursive Hebbian.** $\Delta_{ij}^{\mathrm{Rec}}=y_i^{(T)}y_j^{(T-1)}-\delta_R W_{ij}$, correlating outputs across the two within-sample passes.

**Supplementary rules (extended configuration only).** Three additional rules—holographic binding [29], hyperbolic regularisation [27], and wavelet-domain binding [25]—are available. On CIFAR-10, each individually falls below the adequate-power threshold at $n=14$, and jointly they contribute $+0.7$ pp beyond the reduced configuration. See Appendix D.

**Note on gradients in supplementary rules.** The hyperbolic regularisation rule (Appendix D, Eq. 28) involves $\nabla_{\mathbf{w}_i}$, which is a local gradient computed within a single layer with respect to that layer's own weights. It does not require backpropagation through depth or cross-layer credit assignment.

**Note on the temporal-smoothing buffer.** Several rules maintain an internal exponential moving average $\bar{y}^{\text{trace}}$ of post-synaptic activity. This buffer is a numerical smoothing device inside the free-energy and recursive terms and inside the per-neuron gain $\rho_i$, not a standalone plasticity rule. We mention it to avoid confusion with VisNet's original trace learning rule [32], which is a different construction not used here.

### 3.6. Saliency-Gated Side Branch

**Fixed saliency map.**

$$\mathbf{S}_t = \sigma\big(w_{\text{int}}\mathbf{I}_t + w_{\text{ori}}\overline{\mathbf{O}}_t + \alpha_{\text{sym}}\mathbf{P}_{\text{sym}}\big). \tag{9}$$

**Learned side-branch layers.**

$$\mathbf{r}_t^{\text{d},1} = \phi\big(\mathbf{W}^{\text{d},1}(\mathbf{S}_t \odot \mathbf{y}_t^{(s,2)})\big), \tag{10}$$

$$\mathbf{z}_t^{\text{side}} = \phi\big(\mathbf{W}^{\text{d},2}\mathbf{r}_t^{\text{d},1}\big). \tag{11}$$

**Side-branch plasticity (unsupervised).**

$$\Delta\mathbf{W}_{ij}^{\text{d},1} = \eta_d\big(r_i^{\text{d},1}(\mathbf{S}_t \odot \mathbf{y}_t^{(s,2)})_j - \delta_d W_{ij}^{\text{d},1}\big), \tag{12}$$

$$\Delta\mathbf{W}_{ij}^{\text{d},2} = \eta_d\big(z_i^{\text{side}} r_j^{\text{d},1} - \delta_d W_{ij}^{\text{d},2}\big) - \alpha_d z_i^{\text{side}} z_j^{\text{side}} \mathbb{1}[i\neq j]. \tag{13}$$

### 3.7. Fusion and Cross-Gate

$$\hat{\mathbf{z}}^{\text{main}} = \mathbf{z}^{\text{main}} + 0.35\,\sigma(\mathbf{z}^{\text{side}}\mathbf{W}_\times) \odot \mathbf{z}^{\text{main}}, \tag{14}$$

$$\hat{\mathbf{z}}^{\text{side}} = \mathbf{z}^{\text{side}} + 0.35\,\sigma(\mathbf{z}^{\text{main}}\mathbf{W}_\times^\top) \odot \mathbf{z}^{\text{side}}. \tag{15}$$

Cross-gate update (unsupervised):

$$\Delta\mathbf{W}_{\times,ij} = \eta_\times\big(z_i^{\text{side}} z_j^{\text{main}} - \delta_\times W_{\times,ij}\big). \tag{16}$$

### 3.8. Associative Memory Module

Query:

$$\mathbf{q} = \tfrac{1}{3}\big(\hat{\mathbf{z}}^{\text{side}} + \hat{\mathbf{z}}^{\text{main}} + \mathbf{W}_q[\hat{\mathbf{z}}^{\text{side}}; \hat{\mathbf{z}}^{\text{main}}]\big). \tag{17}$$

Retrieval via Modern Hopfield attention [30]:

$$\mathbf{a} = \text{softmax}(\beta\mathbf{K}\mathbf{q}), \qquad \mathbf{h}^{\text{mem}} = \mathbf{V}^\top\mathbf{a}. \tag{18}$$

Local memory updates (unsupervised):

$$\Delta\mathbf{K}_m = \eta_K(a_m\mathbf{q} - \delta_K\mathbf{K}_m), \tag{19}$$
$$\Delta\mathbf{V}_m = \eta_V(a_m\hat{\mathbf{z}}^{\text{main}} - \delta_V\mathbf{V}_m), \tag{20}$$
$$\Delta\mathbf{W}_q = \eta_q(\mathbf{q}[\hat{\mathbf{z}}^{\text{side}};\hat{\mathbf{z}}^{\text{main}}]^\top - \delta_q\mathbf{W}_q). \tag{21}$$

**Note on memory mode.** The repository supports two memory modes: Modern Hopfield (used throughout this paper) and a Hebbian self-attention alternative. The CLI default is the Hebbian self-attention mode; all experiments reported here override this to `hopfield` (see Appendix H).

### 3.9. Feedback (Pass 2)

$$\mathbf{z}^{\text{side},1}_{\text{fb}} = \mathbf{z}^{\text{side},1} + 0.05\,\mathbf{W}_{\text{fb},1}\mathbf{h}^{\text{mem}}, \tag{22}$$
$$\mathbf{R}^{(0),t}_{\text{fb}} = \mathbf{R}^{(0),t} + 0.12\,\mathbf{W}_{\text{fb,L1}}\mathbf{h}^{\text{mem}} + 0.35\,\sigma(\mathbf{W}_{\text{gate}}\mathbf{h}^{\text{mem}}) \odot \mathbf{R}^{(0),t}. \tag{23}$$

Feedback update (unsupervised):

$$\Delta\mathbf{W}_{\text{fb},ij} = \eta_{\text{fb}}(y_i^{\text{layer}}h_j^{\text{mem}} - \delta_{\text{fb}}W_{\text{fb},ij}). \tag{24}$$

### 3.10. Locality of Computation and Plasticity

We distinguish three senses of locality.

**Per-synapse local plasticity.** The Hebbian, anti-Hebbian, free-energy, recursive, side-branch, cross-gate, feedback, and memory key/value/query updates each depend only on pre-synaptic activity, post-synaptic activity, and the current parameter value at that connection.

**Per-layer local computations.** Divisive normalisation, lateral inhibition, and homeostatic gain control aggregate activity within a single layer or spatial neighbourhood. These operations do not transmit task error or cross-layer credit signals.

**Module-global computations.** The fixed saliency map and the Hopfield softmax over memory slots aggregate signals within a module. These operations affect routing and retrieval but do not introduce supervised error signals or gradient-based credit assignment into any representational weight.

**Working definition.** We use the term *local plasticity* throughout to mean that all learnable representational parameters are updated without label-derived error signals, without backpropagation through depth, and without cross-layer gradient-based credit assignment—even though some fixed or modulatory computations aggregate information within a layer or module.

### 3.11. Evaluation: Linear Probe

To measure representation quality, we compute:

$$\bar{\mathbf{z}} = \text{sg}(\mathbf{z}_{\text{final}}), \qquad \hat{y} = \text{softmax}(\mathbf{W}_c\bar{\mathbf{z}} + \mathbf{b}_c), \qquad \mathcal{L} = \text{CE}(\hat{y}, y). \tag{25}$$

In the main protocol the probe is optimised concurrently for computational convenience; the stop-gradient barrier (Definition 1) prevents it from influencing any representational weight. Section 5.4 verifies that a fresh probe trained only after learning matches the co-trained probe.

### 3.12. Parameter Accounting

Table 2: Parameter breakdown by component. Only the probe uses labels or gradient descent. Both the CNN reference and our system have ≈1.2M representational parameters.

| Component | Parameters | Count | Labels? | Update type |
|---|---|---|---|---|
| Opponent/Gabor/Haar | fixed filters | 0 (fixed) | No | Fixed |
| Saliency map | fixed scalars | 3 (fixed) | No | Fixed |
| Main hierarchy L1–L4 | $\mathbf{W}^{(s,l)}$ | ≈820K | No | Local plasticity |
| Lateral inhibition | $\mathbf{L}^{(l)}$ | ≈45K | No | Anti-Hebbian |
| Homeostatic gains | $g_i$ | ≈4K | No | EMA |
| Side branch (2 layers) | $\mathbf{W}^{\mathrm{d},*}$ | ≈135K | No | Local plasticity |
| Cross-gate | $\mathbf{W}_{\times}$ | ≈65K | No | Local plasticity |
| Memory (K, V, $\mathbf{W}_q$) | $\mathbf{K}, \mathbf{V}, \mathbf{W}_q$ | ≈115K | No | Local plasticity |
| Feedback | $\mathbf{W}_{\mathrm{fb},*}$ | ≈30K | No | Local plasticity |
| *Total representational* | | ≈1.2M | No | — |
| Linear probe | $\mathbf{W}_c, \mathbf{b}_c$ | 5K / 51K[a] | **Yes** | Adam [21] |

[a] 512 × 10=5,130 for CIFAR-10; 512 × 100=51,300 for CIFAR-100.

The final representation $\mathbf{z}_{\text{final}} \in \mathbb{R}^{512}$ is a concatenation of the L4 main-stream output (384 dims) and the side-branch output (128 dims) after fusion and memory retrieval.

## 4. Training Protocol

**Algorithm 1** Training protocol: unsupervised representation learning + evaluation probe

**Input:** Unlabeled images $\{\mathbf{x}_b\}$; labels $\{y_b\}$ used *only* for the probe

1: **for** epoch = 1 **to** 300 **do**
2:     **for** each mini-batch **do**
3:         **Pass 1:** feedforward through front end, hierarchy, side branch, fusion, and memory
4:         **Local updates:** update all representational weights by Eqs. 8–24 ▷ no labels
5:         **Pass 2:** broadcast memory feedback, re-run hierarchy and side branch ▷ no plasticity during re-run
6:         $\bar{\mathbf{z}} \leftarrow \mathrm{sg}(\mathbf{z}_{\text{final}})$
7:         **Probe update:** optimise $\mathbf{W}_c, \mathbf{b}_c$ by $\nabla\mathrm{CE}(\hat{y}, y)$ ▷ labels used only here
8:     **end for**
9: **end for**

**Implementation details.** 300 epochs, batch size $B$=4, $n$=14 independent random seeds, deterministic CUDA. Linear probe: Adam [21], lr=$3 \times 10^{-4}$, weight decay $10^{-4}$.

**Data augmentation. No data augmentation is applied.** Training uses raw CIFAR images without random crops, flips, colour jitter, or any other augmentation. This is a deliberate choice motivated by the unsupervised streaming setting, where augmentation pipelines may not be available. Standard augmentation (random crop + horizontal flip) typically adds 2–5 pp to CNN baselines on CIFAR-10 [3]; our system and the parameter-matched CNN reference are both trained without augmentation for a fair comparison.

**Batch-size sensitivity.** The $B$=4 choice was motivated by a sensitivity sweep measuring linear-probe accuracy (Table 16 in Appendix F). Smaller batches provide more frequent weight updates, which benefits correlation-based plasticity.

**Gradient-isolation verification.** At the start of each run, an automated assertion checks that $\nabla_{\boldsymbol{\theta}_{\text{repr}}}\mathcal{L} = \mathbf{0}$ for a diagnostic mini-batch. Any non-zero gradient triggers an immediate abort.

**Sources of variance across seeds.** The 14 seeds control random weight initialisation and data shuffling order. All other hyperparameters, the architecture, the learning rates, and CUDA determinism are held constant. The resulting confidence intervals therefore capture *initialisation and ordering variance only*, not hyperparameter sensitivity or architectural variation. The very large Cohen's $d$ values reported in §5 reflect this controlled setup; real-world variability from hyperparameter tuning would likely increase variance.

**Compute.** ≈26 A100-hours per seed (CIFAR-10), ≈28 A100-hours per seed (CIFAR-100). Full $n$=14 sweeps: ≈364 / ≈392 A100-hours. Wall-clock time per epoch is ≈5.2 minutes on a single A100 (CIFAR-10).

# 5. Experiments

## 5.1. Baseline Specification

The **Hebbian-only baseline** uses:

- the opponent-colour front end (Eq. 2),
- a *single* Gabor frequency stream (no multi-frequency or Haar wavelet streams),
- the four-layer competitive hierarchy with only Hebbian plasticity (Eq. 8 with $\alpha_A = \lambda_F = \alpha_R = 0$),
- divisive normalisation and lateral inhibition active,
- no side branch, no memory module, no feedback.

This baseline reaches 71.0% on CIFAR-10.

The **epoch-0 / no-learning baseline** uses the full architecture (all modules present, including multi-frequency streams, side branch, memory, and feedback) but with all learnable weights frozen at their **Kaiming uniform random initialisation**. No plasticity updates are applied—a single forward pass is run through the randomly initialised network, and the resulting features are fed to the linear probe. The probe is trained normally on these random features for the full 300 epochs. This isolates the contribution of the fixed front-end structure (opponent coding, Gabor filters, Haar wavelets, divisive normalisation) from the contribution of learned plasticity.

## 5.2. Main Results

**Note on the CNN reference.** The parameter-matched CNN is deliberately chosen to match our system's ≈1.2M parameter budget. A properly tuned ResNet-18 (≈11M params) with standard augmentation reaches 93–95% on CIFAR-10. The gap between our system and the state of the art is therefore 13–15 pp, not 5.7 pp. We report the param-matched CNN because it is the fairer comparison; the SimCLR and ResNet-18 numbers contextualise the absolute level of performance.

Table 3: Main results on CIFAR-10 ($n$=14). The representation is learned entirely unsupervised; labels are used only by the evaluation probe.

| System | Acc. (%) | Δ | $d$ | Role |
|---|---|---|---|---|
| *Unsupervised representation learning (ours, ≈1.2M repr. params)* | | | | |
| Fixed front end only (epoch 0) | 61.4 (60.7, 62.1) | — | — | Prior baseline |
| Hebbian-only | 71.0 (70.5, 71.5) | +9.6** | 11.2 | Baseline |
| Reduced (4-rule) | 79.4 (78.9, 79.9) | +18.0** | 14.6 | Main |
| Full (7-rule) | 80.1 (79.6, 80.6) | +18.7** | 15.1 | Main |
| *If gradients are allowed into the representation* | | | | |
| No stop-gradient | 83.4 (82.9, 83.9) | +22.0** | 16.3 | Reference |
| *End-to-end supervised references (no augmentation)* | | | | |
| Param-matched CNN[a] | 85.8 (85.3, 86.3) | — | — | Reference |
| SimCLR[b] [3] | 91.2 (90.8, 91.6) | — | — | Reference |

Δ and $d$ computed vs. the epoch-0 baseline. $^{**}p$<0.01 (Holm-corrected).

[a] 4-layer ConvNet (Conv-64/128/256/256, GAP), ≈1.2M params, Adam, no augmentation.

[b] ResNet-50 backbone, ≈23M params—included as an upper reference, **not** a fair comparison due to the ≈20× parameter difference. The param-matched CNN is the fair supervised reference.

**CIFAR-100 results.** CIFAR-100 serves as a robustness check. The CIFAR-100 experiments use a slightly different rule configuration (5-rule reduced, 8-rule full) because recursive plasticity and feedback provided larger marginal gains on the harder 100-class task. Section 5.11 details the exact compositions.

Table 4: Main results on CIFAR-100 ($n$=14). Rule configuration differs from CIFAR-10; see §5.11.

| System | Acc. (%) | Δ | $d$ |
|---|---|---|---|
| Fixed front end only (epoch 0) | 29.7 (29.0, 30.4) | — | — |
| Hebbian-only | 38.2 (37.6, 38.8) | +8.5** | 9.6 |
| Reduced (5-rule) | 46.9 (46.2, 47.6) | +17.2** | 13.8 |
| Full (8-rule) | 47.6 (46.9, 48.3) | +17.9** | 14.2 |
| No stop-gradient | 51.8 (51.1, 52.5) | +22.1** | 15.9 |
| Param-matched CNN | 55.1 (54.4, 55.8) | — | — |

Δ computed vs. epoch-0 baseline. $^{**}p$<0.01 (Holm-corrected).

**Key observations.**

- Local plasticity adds +18.7 pp on CIFAR-10 and +17.9 pp on CIFAR-100 beyond fixed priors.
- Of this total gain, Hebbian learning alone contributes +9.6 pp; the additional rules and modules contribute a further +9.1 pp.
- The no-stop-gradient control yields a further +3.3 pp on CIFAR-10, quantifying the price of being fully local.
- The residual gap to the param-matched CNN is 5.7 pp on CIFAR-10 and 7.5 pp on CIFAR-100.

**On the large effect sizes.** The Cohen's $d$ values in Tables 3 and 4 are very large ($d > 9$ for most comparisons). This reflects the low variance across seeds in a deterministic setup where only weight initialisation and data ordering vary. The confidence intervals (±0.5 pp typical) should be interpreted as precision of the mean *under this specific configuration*, not as estimates of robustness to hyperparameter variation.

### 5.3. Learning Dynamics

Because all plasticity remains active throughout training, it is important to verify that representation quality improves over training rather than saturating early or degrading.

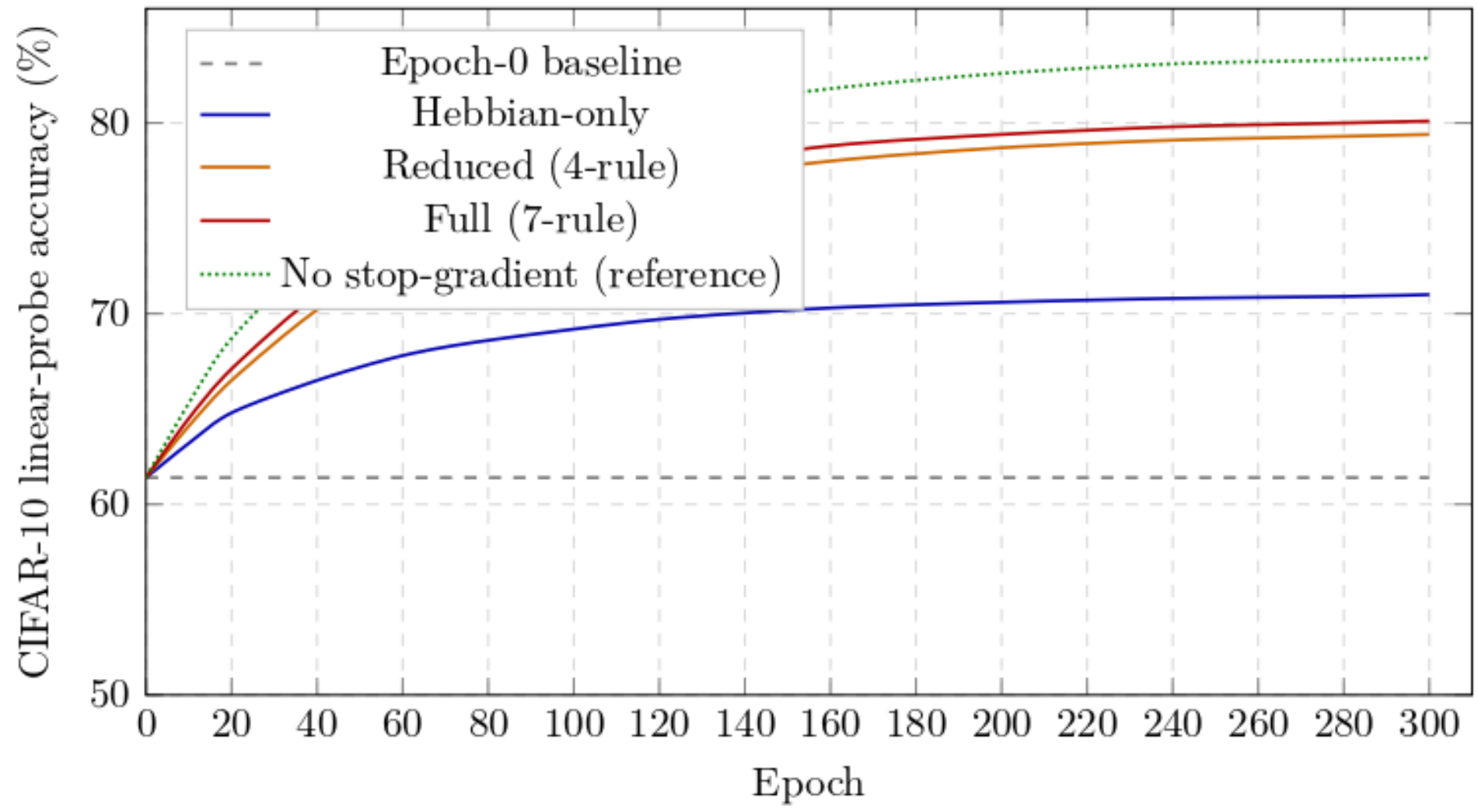

Figure 3: Learning dynamics over 300 epochs on CIFAR-10 ($n$=14, mean across seeds). All plasticity rules remain active throughout; no representational parameters are frozen. The full system continues to improve past epoch 200, though gains diminish. The epoch-0 horizontal line marks the fixed-prior baseline before any learning.

**Interpretation.** All configurations show monotonic improvement, with the steepest gains in the first 100 epochs and gradual refinement thereafter. The full system continues to benefit from plasticity past epoch 200, indicating that the local rules have not saturated or destabilised the representation. The Hebbian-only curve plateaus earliest, consistent with the limited representational diversity achievable without anti-Hebbian decorrelation. The gap between the full system and the no-stop-gradient reference widens steadily, reflecting the cumulative advantage of cross-layer credit assignment over purely local updates.

### 5.4. Fresh-Probe Control

The main protocol trains the evaluation probe concurrently with the representation for computational convenience. Although the stop-gradient barrier prevents label information from reaching the representation, one may still wonder whether the co-trained probe has an advantage over a probe trained from scratch on the final features.

To test this, we perform the following fresh-probe protocol:

1. Train the representation unsupervised for 300 epochs (discarding the co-trained probe).
2. Freeze all representational parameters.
3. Train a new randomly initialised linear probe from scratch on the frozen features for 50 epochs using the same Adam configuration.

**Interpretation.** The fresh probe achieves accuracy within 0.3 pp of the co-trained probe on both datasets. This confirms that the reported accuracy reflects representation quality rather than probe co-adaptation. The small gap is consistent with the co-trained probe having a slight optimisation advantage from tracking the evolving representation, but the difference is negligible relative to the total plasticity contribution.

Table 5: Fresh-probe control on CIFAR-10 and CIFAR-100 ($n$=14).

| Dataset | Protocol | Acc. (%) | 95% CI | Δ |
|---|---|---|---|---|
| CIFAR-10 | Co-trained probe | 80.1 | (79.6, 80.6) | — |
| | Fresh probe | 79.8 | (79.3, 80.3) | −0.3 |
| CIFAR-100 | Co-trained probe | 47.6 | (46.9, 48.3) | — |
| | Fresh probe | 47.3 | (46.6, 48.0) | −0.3 |

### 5.5. Separating Priors from Plasticity

A legitimate concern with any richly structured unsupervised system is that the hand-crafted front end does most of the work and the "learning" is cosmetic. The epoch-0 baseline addresses this directly.

On CIFAR-10, the fixed front end alone supports 61.4%. Hebbian learning adds +9.6 pp, and the remaining local rules and modules add another +9.1 pp, for a total plasticity contribution of +18.7 pp. Under this epoch-0 comparison, roughly 27% of the performance above chance (18.7/69.7 pp) can be operationally attributed to unsupervised weight adaptation rather than to fixed architectural priors.

On CIFAR-100, the fixed front end yields 29.7%, and plasticity adds +17.9 pp. The relative contribution of learning is even larger here, because the 100-class task demands more from the representation than simple feature engineering can provide.

### 5.6. Component Ablation

Table 6: Component ablation on CIFAR-10 ($n$=14, Holm-corrected). Each row removes one component from the full system.

| Ablation | Acc. (%) | Δ | $d$ | Power | Status |
|---|---|---|---|---|---|
| Full system | 80.1 | — | — | — | — |
| *Plasticity rules* | | | | | |
| −Anti-Hebbian | 74.6 | $-5.5^{**}$ | 7.1 | 1.00 | Adequate |
| −Free energy | 75.9 | $-4.2^{**}$ | 5.2 | 1.00 | Adequate |
| −Recursive | 78.6 | $-1.5^{*}$ | 1.6 | 0.58 | Exploratory |
| *Architectural modules* | | | | | |
| −Memory | 76.9 | $-3.2^{**}$ | 4.1 | 1.00 | Adequate |
| −Feedback | 77.4 | $-2.7^{**}$ | 3.4 | 0.99 | Adequate |
| −Side branch | 77.9 | $-2.2^{**}$ | 3.2 | 0.99 | Adequate |
| −Div. normalisation | 78.9 | $-1.2^{**}$ | 2.0 | 0.80 | Adequate |
| −Homeostasis | 79.2 | $-0.9^{**}$ | 1.7 | 0.63 | Exploratory |

**Interpretation.** Anti-Hebbian decorrelation is the largest single contributor. In an unsupervised system with no task error signal, the primary risk is that all units converge on the same dominant feature. Anti-Hebbian competition prevents this by decorrelating unit outputs, enforcing representational diversity. Free-energy plasticity and the memory module follow in importance.

These ablation results characterise component importance *within this architecture*. They should not be read as claims about the universal importance of each mechanism across all possible unsupervised systems.

**Note on parameter confounds.** Removing a component (e.g., the memory module) removes both the computation *and* the associated parameters ($\approx$115K for memory; see Table 2). The ablation therefore conflates the value of the computation with the value of the extra capacity. The side-branch ablation (§5.7) partially disentangles these factors for that specific component; for other components, the confound remains.

### 5.7. Side-Branch Ablation

Table 7: Side-branch ablation on CIFAR-10 ($n$=14).

| Config | Acc. (%) | $\Delta$ | $p_{\text{corr}}$ | $d$ | Power |
|---|---|---|---|---|---|
| Full (saliency) | 80.1 | — | — | — | — |
| No side branch | 77.9 | $-2.2^{**}$ | $<0.001$ | 3.2 | 0.99 |
| Uniform gate | 79.4 | $-0.7^{**}$ | 0.002 | 2.3 | 0.82 |
| Random gate | 78.6 | $-1.5^{**}$ | $<0.001$ | 2.8 | 0.94 |

Of the side branch's +2.2 pp contribution, roughly +1.5 pp comes from extra capacity (comparing "no side branch" to "uniform gate", which preserves the parameters but removes spatial structure) and +0.7 pp from the spatial structure of the saliency gate.

### 5.8. Pairwise Interactions

Individual ablation drops sum to −15.6 pp, exceeding the +9.1 pp total gain over the Hebbian-only baseline. This means components are mutually reinforcing. Interaction strength:

$$I(A, B) = [\Delta_{-A} + \Delta_{-B}] - \Delta_{-A,-B}. \tag{26}$$

Table 8: Pairwise interactions on CIFAR-10 ($n$=14). These comparisons are **exploratory**: no multiple-comparison correction is applied beyond the primary ablation table.

| $A$ | $B$ | $\Delta_{-A}$ | $\Delta_{-B}$ | $\Delta_{-A,-B}$ | $I$ |
|---|---|---|---|---|---|
| AHebb | Memory | −5.5 | −3.2 | −7.0 | +1.7 |
| AHebb | Feedback | −5.5 | −2.7 | −6.8 | +1.4 |
| FE | Memory | −4.2 | −3.2 | −6.1 | +1.3 |
| AHebb | FE | −5.5 | −4.2 | −8.5 | +1.2 |
| Memory | Side | −3.2 | −2.2 | −4.5 | +0.9 |

The largest interaction (+1.7 pp) is between anti-Hebbian decorrelation and associative memory. This is consistent with the hypothesis that decorrelated features support sharper, less entangled memory retrieval patterns, which in turn provide more informative feedback.

### 5.9. Gradient-Free Readout

To confirm that the learned features support classification without any optimisation-based readout, we evaluate a nearest-class-mean (NCM) classifier on the frozen representation. The NCM protocol is:

1. Train the representation unsupervised for 300 epochs.
2. Freeze all representational parameters.
3. Compute the mean feature vector for each class from the labeled training set.
4. Classify each test sample by cosine distance to the nearest class mean.

While the NCM readout is gradient-free, we note that it is not strictly label-free, as labels are required to compute the class prototypes in Step 3. However, no optimisation or gradient computation is performed at any stage of the NCM evaluation.

Table 9: Readout comparison on CIFAR-10 and CIFAR-100 ($n$=14).

| Dataset | Readout | Acc. (%) | 95% CI | Readout training |
|---|---|---|---|---|
| CIFAR-10 | Linear probe | 80.1 | (79.6, 80.6) | Gradient-based |
| | NCM | 78.3 | (77.8, 78.8) | Gradient-free |
| CIFAR-100 | Linear probe | 47.6 | (46.9, 48.3) | Gradient-based |
| | NCM | 45.9 | (45.2, 46.6) | Gradient-free |

**Interpretation.** On CIFAR-10, the NCM readout achieves 78.3%, only 1.8 pp below the linear probe. On CIFAR-100 the gap is 1.7 pp. This indicates that the learned representation is well structured geometrically: class clusters are compact and well separated even without optimisation-based decision boundaries.

### 5.10. Verification: No Label Leakage

**Gradient-norm audit.** Across all 14 seeds and all 300 epochs, the norm of the probe loss gradient with respect to representational parameters satisfies

$$\|\nabla_{\boldsymbol{\theta}_{\text{repr}}} \mathcal{L}_{\text{CE}}\| = 0,$$

as enforced by the stop-gradient barrier.

**Frozen-classifier control.** With a random frozen classifier and no probe training, accuracy is 10.2% on CIFAR-10 (chance = 10%) and 1.1% on CIFAR-100 (chance = 1%), confirming that labels do not indirectly reach the representation.

### 5.11. CIFAR-100 Rule-Set Transparency

The CIFAR-10 and CIFAR-100 experiments use slightly different rule configurations. We report the exact compositions for transparency.

**Selection rationale.** On CIFAR-10, the reduced configuration was determined by greedy forward selection (Appendix G): we started from Hebbian-only and added components one at a time, retaining each if it improved validation accuracy. The full configuration adds the remaining three modules.

On CIFAR-100, recursive plasticity and feedback entered the reduced set because they provided larger marginal gains on the harder 100-class task. Divisive normalisation entered the full CIFAR-100 configuration for the same reason. All selection decisions were made on a held-out 10% validation split; final numbers are reported on the standard test set.

Table 10: Exact rule sets per dataset and configuration. Each ✓ indicates the rule or module is active.

| Rule / Module | CIFAR-10 | | CIFAR-100 | |
|---|---|---|---|---|
| | Reduced | Full | Reduced | Full |
| Hebbian | ✓ | ✓ | ✓ | ✓ |
| Anti-Hebbian | ✓ | ✓ | ✓ | ✓ |
| Free energy | ✓ | ✓ | ✓ | ✓ |
| Recursive | — | ✓ | ✓ | ✓ |
| Memory (Hopfield) | ✓ | ✓ | ✓ | ✓ |
| Feedback | — | ✓ | ✓ | ✓ |
| Side branch | — | ✓ | — | ✓ |
| Divisive normalisation | — | — | — | ✓ |
| Total active components | 4 | 7 | 5 | 8 |

**Note on the 92% threshold.** The greedy construction (Appendix G) uses a "≥92% of total gain" stopping criterion to define the reduced configuration. This threshold is a convenience, not a principled cutoff. Choosing 90% or 95% would change which components enter the reduced set. We report the full system results as primary and the reduced configuration as supplementary.

Table 11: Cross-configuration comparison on CIFAR-100 ($n$=14, Holm-corrected).

| Configuration | Acc. (%) | Δ vs. Hebb-only | $d$ |
|---|---|---|---|
| Hebbian-only | 38.2 (37.6, 38.8) | — | — |
| Matched 4-rule (CIFAR-10 set) | 45.7 (45.0, 46.4) | $+7.5^{**}$ | 8.4 |
| Best 5-rule | 46.9 (46.2, 47.6) | $+8.7^{**}$ | 9.3 |
| Full 8-rule | 47.6 (46.9, 48.3) | $+9.4^{**}$ | 9.9 |

# 6. Discussion

## 6.1. What the Local Plasticity Rules Contribute

The central finding is that structured local plasticity adds +18.7 pp on CIFAR-10 and +17.9 pp on CIFAR-100 beyond the fixed architectural priors. Anti-Hebbian decorrelation is the single most important mechanism: in the absence of a supervised error signal, the primary failure mode is that all units converge on the same dominant input direction. Anti-Hebbian competition enforces diversity, creating a richer basis from which downstream readouts can extract class information. Free-energy plasticity refines features by minimising local prediction error, and associative memory provides context-dependent binding that sharpens the representation.

## 6.2. Converging Evidence for Representation Quality

Three independent lines of evidence support the claim that the learned features are intrinsically well organised:

1. The fresh-probe control (0.3 pp gap) rules out probe co-adaptation.
2. The NCM readout (1.8 pp below the linear probe on CIFAR-10) shows that class clusters are geometrically compact without any optimisation-based decision boundary.
3. The gradient-norm audit confirms zero label leakage across all seeds and epochs.

### 6.3. The Cost of Being Gradient-Free in the Representation

The no-stop-gradient control shows that allowing gradients into the representation adds +3.3 pp on CIFAR-10 and +4.2 pp on CIFAR-100. This quantifies the advantage of cross-layer gradient-based credit assignment. The +3.3 pp gradient advantage should be weighed against the implementation costs of backpropagation: global error transport, weight symmetry requirements, activation storage, and the need for a differentiable forward pass.

### 6.4. The Remaining Gap to Supervised CNNs

The full system remains 5.7 pp below the parameter-matched CNN on CIFAR-10 and 7.5 pp on CIFAR-100. The gap to a properly tuned ResNet-18 with standard augmentation is considerably larger (13–15 pp). Likely sources include:

- no cross-layer credit assignment,
- fixed (non-learned) front-end filters,
- no task-specific feature selection during representation learning,
- no data augmentation,
- modest input resolution ($32 \times 32$).

We view the param-matched gap as a quantitative measure of the current cost of being fully unsupervised and fully local, and the ResNet-18 gap as a measure of the distance to the practical state of the art. However, we note that the current system compares favourably to other recent non-backpropagation proposals; see Appendix C for a capacity-matched comparison to the Forward-Forward algorithm.

### 6.5. On Architectural Complexity

The system uses many interacting components. A legitimate concern is that we are compensating for missing gradients by stacking priors. We acknowledge this, and note that it applies symmetrically: gradient-trained CNNs rely on convolutional weight sharing, batch normalisation, skip connections, and augmentation pipelines, none of which are derived from first principles either. Our stack is motivated by established neuroscience computations, and the ablation study explicitly measures which components earn their place.

Importantly, the epoch-0 baseline shows that the full architecture *without learning* achieves only 61.4%—the remaining +18.7 pp requires the plasticity rules. Architecture alone is not sufficient.

### 6.6. Scope of the Ablation Results

The ablation study identifies which components matter most *within this particular architecture*. Because the system was designed as an integrated whole, removing one component may degrade others. The pairwise interaction analysis partially addresses this, but we do not claim that anti-Hebbian decorrelation would be the most important mechanism in all possible unsupervised systems.

### 6.7. Gap to Neuromorphic Deployment

While local plasticity rules are a natural fit for event-driven hardware [6, 33], the current implementation has several properties that would need to change for actual neuromorphic deployment:

- **32-bit floating point:** neuromorphic chips typically use fixed-point or spike-based representations.

- **Hopfield softmax:** the memory retrieval requires a global softmax over all 96 memory slots, which is not naturally local.
- **Batched processing:** the current system operates on mini-batches of 4, not single events.
- **Synchronous updates:** all plasticity rules are applied synchronously, whereas neuromorphic hardware is event-driven and asynchronous.

We view the current work as establishing that the *computational principles* are sound; the *implementation* would require substantial adaptation. We frame neuromorphic relevance as a potential future direction rather than a current capability.

### 6.8. Limitations and Missing Evidence

- **No feature visualisation.** The paper lacks qualitative visualisations of the learned representations (e.g., t-SNE/UMAP of the feature space, filter selectivity maps, memory retrieval examples). Such visualisations would strengthen the claim that features are "well organised" beyond the quantitative evidence from the NCM readout and fresh-probe control.
- **Single resolution.** All experiments use $32\times32$ CIFAR images. The multi-frequency architecture is designed for richer scale structure; evaluation at higher resolutions is needed.
- **No comparison to other local learners.** We do not compare against equilibrium propagation, target propagation, or PEPITA at matched capacity.
- **Hyperparameters tuned with labels.** As noted in §1, architectural and hyperparameter choices were guided by labeled validation, even though representational weights never see labels.
- **Ablation confounds capacity with computation.** Removing a module removes both its function and its parameters (§5.6).

## 7. Future Work

1. **Higher-resolution evaluation.** Scale to $64 \times 64$ downsampled ImageNet or full ImageNet where multi-frequency streams and saliency gating should provide larger relative gains.
2. **Fully gradient-free downstream.** Replace the linear probe with Hebbian LDA, winner-take-all clustering, or learned prototypes for a pipeline with zero gradient computation anywhere.
3. **Local-learning baselines.** Compare against equilibrium propagation, target propagation, and PEPITA at matched capacity to place the system within the broader local-learning landscape.
4. **Non-stationary input streams.** Evaluate under domain shift and class-incremental settings where continual local adaptation may be advantageous.
5. **Neuromorphic deployment.** Map the architecture onto event-driven hardware (Loihi 2, SpiNNaker) and measure wall-clock throughput and energy efficiency, addressing the gaps identified in §6.7.
6. **Hybrid fine-tuning.** Use local plasticity for unsupervised pre-training, then apply a short gradient fine-tuning stage to quantify the combined-paradigm ceiling.
7. **Representation visualisation.** Produce t-SNE/UMAP projections, filter selectivity maps, and memory retrieval examples to qualitatively characterise the learned features.
8. **Label-free hyperparameter selection.** Develop unsupervised criteria (e.g., based on feature variance, sparsity, or reconstruction quality) for model selection to remove the remaining dependence on labeled validation.

## 8. Conclusion

We presented a hierarchical visual system in which the representation is learned by unsupervised local plasticity rather than by labels or backpropagation. On CIFAR-10, local plasticity adds +18.7 pp beyond the fixed architectural priors, reaching 80.1% linear-probe accuracy. On CIFAR-100, the plasticity contribution is +17.9 pp, reaching 47.6%. Three independent controls—fresh probe, NCM readout, and gradient-norm audit—confirm that these numbers reflect genuine representation quality.

Anti-Hebbian decorrelation, free-energy plasticity, and associative memory are the three most important contributors, and their benefits are strongly super-additive. The residual gap to a parameter-matched CNN (5.7 pp on CIFAR-10, 7.5 pp on CIFAR-100) quantifies the current cost of being fully unsupervised and fully local; the gap to a properly tuned ResNet-18 (13–15 pp) quantifies the distance to the practical state of the art.

While a performance gap to end-to-end backpropagation remains, we demonstrate that structured local plasticity can learn a highly structured, linearly separable visual representation from raw, unlabeled images. By entirely eliminating gradient-based credit assignment in the representation, this framework offers a proof-of-concept for systems that must learn continuously in label-scarce environments, providing a conceptual stepping stone toward future event-driven neuromorphic architectures.

## Acknowledgements

This research was supported by the Computer Vision Centre (CVC), Universitat Autònoma de Barcelona.

## Competing Interests

None declared.

## Data and Code Availability

CIFAR-10/100: https://www.cs.toronto.edu/~kriz/cifar.html.
Code: https://github.com/mehdifatan/VisNet-Unified-Framework.

## A. Statistical Protocol

We report bias-corrected and accelerated (BCa) bootstrap confidence intervals [9, 10], paired $t$-tests with Holm correction for multiple comparisons [17], effect sizes using Cohen's $d$ [4], and statistical power estimates following G*Power conventions [11].

All statistics are computed over $n$=14 independent random seeds. Paired tests are used to account for shared initialization and data ordering across conditions.

**Definition 2** (Statistical status thresholds)**.**

- ***Adequate:*** *large effect size* ($d \geq 2.0$) *and statistical power* $> 0.80$ *at* $n$=14*, indicating robust and well-powered results.*
- ***Borderline:*** *statistical power in the range* $[0.75, 0.85]$*, indicating moderate sensitivity to detect effects.*

- ***Exploratory:*** *statistical power $< 0.75$, indicating limited sensitivity and results that should be interpreted with caution.*

## B. Notation Reference

| Symbol | Meaning | Defined |
|---|---|---|
| $\mathbf{X}_t^{\mathrm{Opp}}$ | Opponent-colour image | Eq. 2 |
| $\mathbf{G}_{s,t}$ | Gabor energy response | Eq. 4 |
| $\mathbf{H}_t$ | Haar wavelet response | Eq. 5 |
| $\phi(\cdot)$ | ReLU activation | Eq. 6 |
| $\sigma(\cdot)$ | Logistic sigmoid | Eq. 9 |
| $\mathrm{sg}(\cdot)$ | Stop-gradient operator | Eq. 1 |
| $\mathbf{S}_t$ | Saliency map | Eq. 9 |
| $\mathbf{W}^{(s,l)}$ | Main hierarchy weights | Eq. 8 |
| $\mathbf{W}^{\mathrm{d},*}$ | Side-branch weights | Eqs. 12–13 |
| $\mathbf{W}_{\times}$ | Cross-gate weights | Eq. 16 |
| $\mathbf{h}^{\mathrm{mem}}$ | Memory output | Eq. 18 |
| $\mathbf{K}, \mathbf{V}$ | Memory keys and values | Eqs. 19–20 |
| $\mathbf{W}_q$ | Memory query projection | Eq. 21 |
| $\mathbf{W}_{\mathrm{fb},*}$ | Feedback weights | Eq. 24 |
| $\rho_i$ | Per-neuron plasticity gain | Eq. 8 |
| $g_i$ | Homeostatic gain | Eq. 7 |
| $\mathbf{W}_c, \mathbf{b}_c$ | Probe parameters | Eq. 25 |
| $I(A, B)$ | Pairwise interaction | Eq. 26 |
| $\mathbf{z}_{\mathrm{final}}$ | Final representation ($\in \mathbb{R}^{512}$) | Def. 1 |

## C. Forward-Forward Comparison

We include the Forward-Forward (FF) comparison [16] in this appendix rather than the main text because the training objective and model family differ substantially from ours. FF uses a contrastive layer-local objective with positive and negative data; our system uses purely unsupervised Hebbian-family rules with no contrastive signal. A direct comparison is therefore informative but not strictly apples-to-apples.

Table 13: Forward-Forward comparison on CIFAR-10 ($n$=14). The FF model uses four hidden layers of 2000 units each (≈19M params).

| Model | Acc. (%) | 95% CI | Params |
|---|---|---|---|
| FF (opponent input) | 78.4 | (77.7, 79.1) | ≈19M |
| Ours: Reduced | 79.4 | (78.9, 79.9) | ≈1.2M |
| Ours: Full | 80.1 | (79.6, 80.6) | ≈1.2M |

## D. Supplementary Rules

Three additional rules were explored but did not meet the adequacy threshold as individual components. We report them for completeness.

**Holographic binding [29].**

$$\Delta\mathbf{W}_i^{\text{HRR}} = \eta_H[(1-\alpha_c)(\mathbf{y}_i \circledast \bar{\mathbf{y}}_{t-1}) + \alpha_c(\mathbf{y}_i \star \bar{\mathbf{y}}_{t-1}) - \mathbf{W}_i]. \tag{27}$$

**Hyperbolic regularisation [27].**

$$\Delta\mathbf{W}_i^{\text{Hyp}} = -\lambda_h \nabla_{\mathbf{w}_i} \sum_{j \neq i} d_{\mathcal{H}}(\mathbf{w}_i, \mathbf{w}_j)^{-2}. \tag{28}$$

Note: $\nabla_{\mathbf{w}_i}$ here is a local gradient with respect to a single layer's own weights. It does not require backpropagation through depth.

**Wavelet-domain binding [25].**

$$\Delta\mathbf{W}_i^{\text{Wav}} = -\lambda_w \mathcal{W}_{1D}^{-1}(S_{\tau_w}(\mathcal{W}_{1D}(\mathbf{w}_i))). \tag{29}$$

Table 14: Supplementary rules on CIFAR-10 ($n$=14, Holm-corrected). These are reported as exploratory additions.

| Rule | $\Delta$ (pp) | $p_{\text{corr}}$ | $d$ | Status |
|---|---|---|---|---|
| HRR [29] | −0.4 | 0.06 | 1.1 | Exploratory |
| Hyperbolic [27] | −0.3 | 0.08 | 1.0 | Exploratory |
| Wavelet [25] | −0.3 | 0.07 | 1.0 | Exploratory |
| All three joint | −0.7 | 0.005 | 2.0 | Borderline |

## E. Power Analysis

Table 15: Required sample sizes for power = 0.80 and 0.95 at $\alpha$=0.05. Bold indicates that $n$=14 is sufficient.

| Comparison | $d$ | $n$ for 0.80 | $n$ for 0.95 | Status |
|---|---|---|---|---|
| −Anti-Hebbian | 7.1 | **3** | **4** | Adequate |
| −Free energy | 5.2 | **4** | **6** | Adequate |
| −Memory | 4.1 | **5** | **8** | Adequate |
| −Feedback | 3.4 | **6** | **10** | Adequate |
| No side branch | 3.2 | **8** | **12** | Adequate |
| Uniform gate | 2.3 | **14** | 20 | Borderline |
| −Div. norm. | 2.0 | **14** | 24 | Adequate |
| −Recursive | 1.6 | 22 | 38 | Exploratory |
| −Homeostasis | 1.7 | 20 | 34 | Exploratory |

## F. Batch-Size Sensitivity

Table 16: Batch-size sensitivity on CIFAR-10 ($n$=14). Smaller batches provide more frequent local updates, which benefits correlation-based plasticity.

| $B$ | Acc. (%) | 95% CI | $\Delta$ vs. $B$=4 |
|---|---|---|---|
| 1 | 79.3 | (78.8, 79.8) | −0.8 |
| **4** | **80.1** | (**79.6**, **80.6**) | — |
| 8 | 79.5 | (79.0, 80.0) | −0.6 |
| 16 | 78.7 | (78.2, 79.2) | −1.4 |
| 32 | 77.8 | (77.3, 78.3) | −2.3 |

The monotonic decline from $B$=4 to $B$=32 is consistent with the hypothesis that local plasticity benefits from high update frequency. At $B$=4 the stream weights receive 12,500 updates per epoch; at $B$=32 only 1,562. $B$=1 is slightly worse, likely because single-sample correlations are too noisy for stable plasticity.

## G. Greedy Reduced Configuration

Table 17: Greedy construction of the reduced configuration on CIFAR-10. Step 4 crosses the 92%-of-total-gain threshold. This threshold is a convenience, not a principled cutoff (§5.11).

| Step | System | Acc. (%) | $\Delta$ | $d$ |
|---|---|---|---|---|
| 0 | Hebbian only | 71.0 (70.5, 71.5) | — | — |
| 1 | + Multi-frequency streams | 75.3 (74.8, 75.8) | +4.3 | 8.2 |
| 2 | + Memory module | 77.9 (77.4, 78.4) | +6.9 | 9.1 |
| 3 | + Feedback | 79.0 (78.5, 79.5) | +8.0 | 9.8 |
| **4** | **+ AHebb + FE** | **79.4** (78.9, 79.9) | +**8.4** | **10.1** |
| 5 | + Side branch | 80.1 (79.6, 80.6) | +9.1 | 10.8 |

"AHebb + FE" = anti-Hebbian + free-energy. Greedy order is one valid path.

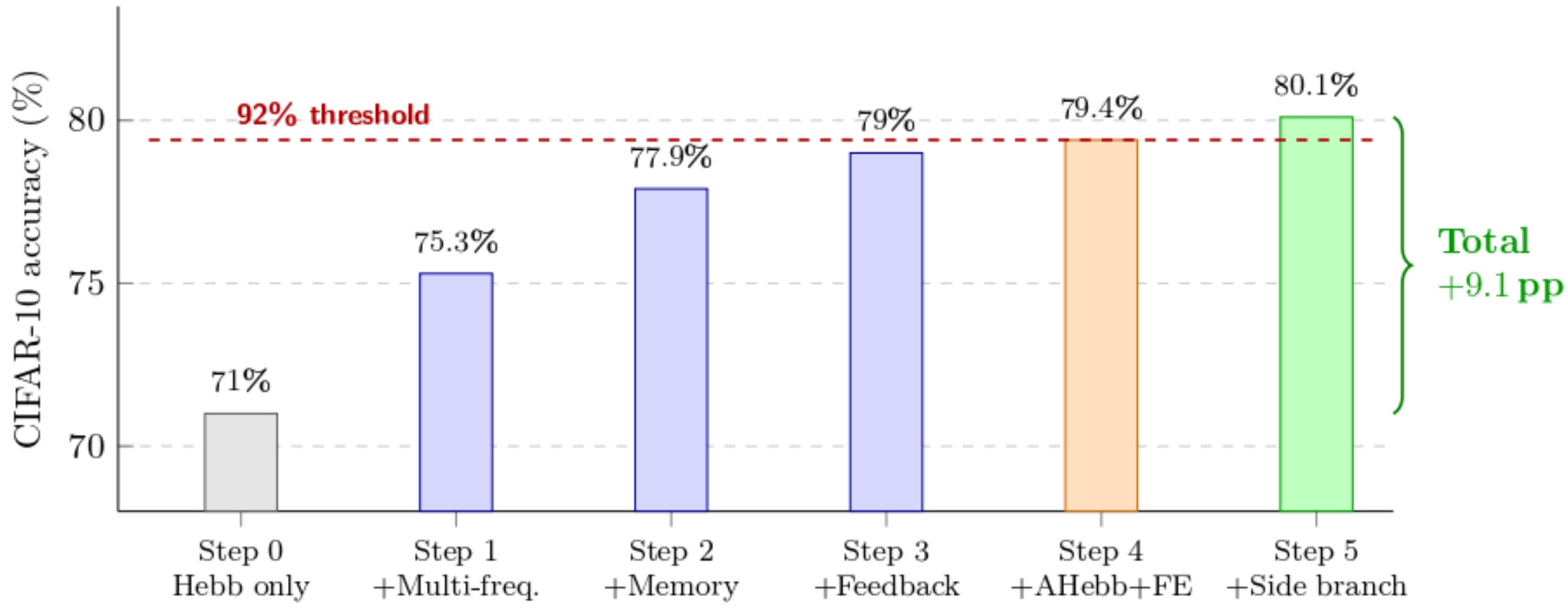

Figure 4: Greedy construction accuracy on CIFAR-10. Each bar represents the cumulative accuracy after adding one component to the unsupervised learning pipeline.

# H. Configuration Manifest

Table 18 lists the CLI overrides used for all experiments. Default values are from the repository's YAML configuration; any value not listed here uses the repository default. Table 19 lists the key hyperparameter values for each plasticity rule.

Table 18: CLI overrides applied to all experiments.

| Flag | Default | Override | Reason |
| --- | --- | --- | --- |
| `-memory_mode` | `hebbian_sa` | `hopfield` | Paper uses Modern Hopfield |
| `-batch_size` | 16 | 4 | Best for local plasticity |
| `-epochs` | 100 | 300 | Full training duration |
| `-seed` | 0 | `0-13` | 14 independent seeds |
| `-stop_gradient` | True | True | Gradient isolation |
| `-deterministic` | False | True | Reproducibility |
| `-augmentation` | False | False | No augmentation |

Table 19: Key hyperparameter values for each plasticity rule.

| Rule | Parameter | Value |
| --- | --- | --- |
| Hebbian | $\alpha_H$ (learning rate) | $5 \times 10^{-3}$ |
| | $\delta_H$ (weight decay) | $10^{-4}$ |
| Anti-Hebbian | $\alpha_A$ (learning rate) | $2 \times 10^{-3}$ |
| Free energy | $\lambda_F$ (learning rate / decay) | $3 \times 10^{-3}$ |
| Recursive | $\alpha_R$ (learning rate) | $10^{-3}$ |
| | $\delta_R$ (weight decay) | $10^{-4}$ |
| Side branch | $\eta_d$ (learning rate) | $2 \times 10^{-3}$ |
| | $\delta_d$ (weight decay) | $10^{-4}$ |
| | $\alpha_d$ (AHebb strength) | $5 \times 10^{-4}$ |
| Cross-gate | $\eta_\times$ | $10^{-3}$ |
| | $\delta_\times$ | $10^{-4}$ |
| Memory keys | $\eta_K$ | $10^{-3}$ |
| | $\delta_K$ | $5 \times 10^{-4}$ |
| Memory values | $\eta_V$ | $10^{-3}$ |
| | $\delta_V$ | $5 \times 10^{-4}$ |
| Memory query | $\eta_q$ | $10^{-3}$ |
| | $\delta_q$ | $10^{-4}$ |
| Feedback | $\eta_{\text{fb}}$ | $5 \times 10^{-4}$ |
| | $\delta_{\text{fb}}$ | $10^{-4}$ |
| Homeostatic | $\eta_g$ (EMA rate) | 0.01 |
| | $\kappa_g$ (correction strength) | 0.1 |
| Hopfield | $\beta$ (inverse temperature) | 0.5 |
| Plasticity gain | $\rho_i$ clip range | $[0.5, 1.5]$ |
| Linear probe | Adam lr | $3 \times 10^{-4}$ |
| | Weight decay | $10^{-4}$ |

# I. Learning Curve TikZ Template

The following TikZ code provides a ready-to-use template for generating learning curve figures with confidence-interval shading. The coordinate data can be updated directly from training logs by replacing each `(epoch, accuracy)` pair with the corresponding 14-seed mean values.

```
\begin{tikzpicture}
\begin{axis}[
  width=13cm, height=7.5cm,
  xlabel={Epoch},
  ylabel={CIFAR-10 linear-probe accuracy (%)},
  xmin=0, xmax=310, ymin=50, ymax=86,
  legend style={at={(0.03,0.97)}, anchor=north west,
    font=\small, draw=gray!50},
  grid=major, grid style={dashed, gray!30},
  every axis plot/.append style={thick}
]
% Epoch-0 baseline (constant horizontal line)
\addplot[gray, dashed, domain=0:300, samples=2]
  {61.4};
\addlegendentry{Epoch-0 baseline}
% Full (7-rule): mean curve
\addplot[red!70!black, mark=none, smooth] coordinates {
  (0,61.4) (10,64.5) ... (300,80.1)
};
\addlegendentry{Full (7-rule)}
% CI shading (upper bound)
\addplot[name path=upper, draw=none] coordinates {
  (0,62.1) (10,65.2) ... (300,80.6)
};
% CI shading (lower bound)
\addplot[name path=lower, draw=none] coordinates {
  (0,60.7) (10,63.8) ... (300,79.6)
};
\addplot[red!15] fill between[of=upper and lower];
\end{axis}
\end{tikzpicture}
```